\documentclass[journal]{IEEEtran}
\IEEEoverridecommandlockouts

\usepackage{cite}
\usepackage{amsmath,amssymb,amsfonts}
\usepackage{graphicx}
\usepackage{textcomp}
\usepackage{xcolor}
\usepackage{booktabs}
\usepackage{multirow}
\usepackage{stfloats}
\usepackage{url}
\usepackage[
    colorlinks=true,
    linkcolor=black,
    citecolor=black,
    urlcolor=blue,
    breaklinks=true
]{hyperref}

\def\BibTeX{{\rm B\kern-.05em{\sc i\kern-.025em b}\kern-.08em
    T\kern-.1667em\lower.7ex\hbox{E}\kern-.125emX}}

\begin{document}

\title{Under-Mattress Temporal Sensing for Next-Day Agitation Risk Scoring in Dementia Wards}

\author{%
\IEEEauthorblockN{%
Zhen Liu\textsuperscript{1},
Marta Bono\textsuperscript{1},
Robbe Decloedt\textsuperscript{2},
Ajda Flisar\textsuperscript{2},
\\
Maarten Van Den Bossche\textsuperscript{2,3,5,$\dagger$},
Maarten De Vos\textsuperscript{1,4,$\dagger$}%
\\}
\IEEEauthorblockA{%
\textsuperscript{1}STADIUS Center, Department of Electrical Engineering, KU Leuven, Leuven, Belgium\\
\textsuperscript{2}Geriatric Psychiatry, University Psychiatric Center KU Leuven, Leuven, Belgium\\
\textsuperscript{3}Neuropsychiatry, Department of Neurosciences, Leuven Brain Institute, 
KU Leuven, Leuven, Belgium\\
\textsuperscript{4}Woman and Child, Department of Development and Regeneration, KU Leuven, Leuven, Belgium\\
\textsuperscript{5}Department of Psychiatry, Maastricht University Medical Centre+ (MUMC+), Maastricht, The Netherlands\\
\textsuperscript{$\dagger$}Both senior authors contributed equally to this study%
}
}

\maketitle

\thispagestyle{plain}
\pagestyle{plain}

\begin{abstract}
\textit{Objective:} Agitation fluctuates over short time horizons in people living with dementia, yet continuous physiological information for anticipating next-day risk is limited. We assessed whether contactless under-mattress signals from the preceding night inform next-day agitation risk and whether preserving minute-level temporal structure improves performance over conventional nightly summaries. \textit{Methods:} We analyzed 423 patient-nights from 65 subjects in a specialized hospital dementia unit using two under-mattress sensing systems. A unified four-paradigm benchmark compared nightly handcrafted summaries, three-period handcrafted features, full-night sequence modeling, and sliding-window multiple-instance learning. Source-specific preprocessing and five-fold patient-grouped cross-validation were used, with performance estimated from pooled out-of-fold predictions. Evaluation included discrimination, calibration, fixed-threshold metrics, and a comparison of period--signal attribution patterns across two temporal models. \textit{Results:} Full-night sequence modeling achieved the highest discrimination (AUROC, 0.692; AUPRC, 0.849) and balanced accuracy (0.658). Both minute-level pipelines had higher AUROC than nightly summaries, but differences from three-period handcrafted features were uncertain. Cross-model attribution prioritized activity, heart rate, and respiratory rate during the core overnight period. Calibration remained limited. \textit{Conclusion:} The preceding night's signals supported modest next-day risk discrimination, with minute-level temporal modeling outperforming nightly summaries. Prospective calibration and external validation are needed before use in individual care decisions. \textit{Significance:} This patient-grouped benchmark identifies contactless overnight sensing as a promising biomedical engineering direction for agitation-risk research in hospitalized dementia cohorts.
\end{abstract}

\begin{IEEEkeywords}
Dementia, Alzheimer’s disease, agitation, sleep physiological signals, ballistocardiography, machine learning, multiple-instance learning, temporal representation, risk scoring.
\end{IEEEkeywords}

\section{Introduction}
\label{sec:intro}
Agitation is one of the most prevalent and clinically burdensome neuropsychiatric symptoms in people living with dementia (PLWD)~\cite{cerejeira2012bpsd}. It includes excessive motor activity, verbal agitation, and aggressive behavior associated with emotional distress~\cite{sano2024agitation}. In care homes, clinically significant agitation affects approximately 40\% of residents with dementia~\cite{livingston2017marque}. Because pharmacological management has limited efficacy and substantial safety concerns~\cite{carrarini2021agit_treat,watt2026management,schneider2005med_risk}, earlier identification of elevated risk could support proactive non-pharmacological care. Agitation varies over short time horizons, while routine clinical observations provide limited continuous information for estimating risk on the following day.

Night-time physiology is a plausible source of short-horizon agitation information, but its predictive value remains under-specified. Sleep disturbances are common in dementia and have been associated with next-day behavioral symptoms~\cite{shi2018sleepDementiaReview,webster2020sleepPrevalenceReview,brown2015sleep,georgescu2026agitation}. More specific longitudinal evidence suggests that sleep--agitation coupling varies across individuals~\cite{single2022AgitActig}, that sleep patterns and behavioral symptoms show day-to-day temporal interdependence~\cite{cho2026sleepBpsd}, and that long-term heart rate variability measures are associated with agitation risk in Alzheimer's disease~\cite{liu2023ANSAgit}. These findings motivate testing whether nocturnal cardiorespiratory and activity signals recorded during the preceding night are associated with agitation risk on the following day in dementia care.

Contactless under-mattress sensing offers a practical route to measure these signals without adding burden to PLWD or ward staff. Commercially available devices, including EMFIT QS\textsuperscript{\textregistered} (EMFIT) and Withings Sleep Analyzer\textsuperscript{\textregistered} (WSA), have been used in sleep and digital health studies~\cite{kholghi2022emfitValidation,ravindran2023contactless,soreq2025WSAsleepPatternAD,ravindran2024reliable,bafaloukou2025interpretable}. These two sensors use pressure-sensitive under-mattress sensing to capture body movement and respiratory signals and derive cardiac measures from ballistocardiographic (BCG) signals. Independent evaluations against polysomnography and actigraphy support their utility for low-burden longitudinal monitoring of sleep physiology, particularly cardiorespiratory and movement-derived signals~\cite{ravindran2023contactless,ravindran2024reliable}. Unobtrusive mattress-based BCG has also been used to estimate nocturnal awakening and sleep efficiency~\cite{jung2014nocturnal}, while contactless radar has recovered cardiorespiratory, movement, and sleep-stage information in older people and people with neurodegenerative disorders~\cite{yin2025unobtrusive}. Their relevance to agitation risk scoring lies in the within-night physiological trajectories they provide during routine overnight care.

Existing sensor-based studies have not established whether physiological signals recorded during the preceding night can predict agitation on the following day in a specialized hospital dementia unit. Prior studies have focused on home monitoring, multimodal sensing, or measurements aggregated over several days~\cite{husebo2020ReviewAgiSensor,deters2024ReviewAgiSensor}, which represent different prediction settings and time horizons. Among the most closely related predictive studies, Ramesh et al.~\cite{ramesh2025} selected LightGBM for next-day prediction from daily aggregated sleep and physiological variables, whereas Bafaloukou et al.~\cite{bafaloukou2025interpretable} selected LightGBM for agitation monitoring over 8-day periods using multimodal in-home variables. These studies compress sensor streams into engineered summary features or aggregate exposure variables, including minimum respiratory rate, mean heart rate, respiratory-rate variability, and awake ratio~\cite{bafaloukou2025interpretable,ramesh2025}. Their findings demonstrate the predictive utility of engineered summaries but leave unclear whether within-night temporal structure adds information beyond nightly summaries. Sequence-aware models have exploited neighboring and longer-range temporal context in automatic sleep staging~\cite{phan2019joint,phan2022sleeptransformer}, while multiple-instance learning (MIL) has been applied to long-term physiological recordings~\cite{ilse2018attention,han2023mamil}. However, whether fine-grained nocturnal structure supports prediction of a distinct next-day behavioral endpoint remains unclear in small repeated-measures hospital cohorts, where evaluation must account for event prevalence, threshold-dependent operating characteristics, device-source heterogeneity, and between-patient differences.

A related explanatory study analyzed an earlier subset from the same hospital monitoring program using mixed-effects models to examine associations between handcrafted nocturnal features and agitation occurrence and severity while accounting for repeated observations~\cite{liu2026dissociating}. The present study addresses a distinct predictive question by evaluating held-out-patient next-day risk scoring and comparing engineered and time-series representations.

Accordingly, this study evaluates whether under-mattress signals recorded during the preceding night support cross-patient next-day agitation risk estimation. We compared four pipelines that represented the preceding night as whole-night handcrafted summaries, period-resolved handcrafted features, a complete minute-level sequence, or overlapping windows with MIL aggregation. Performance was evaluated using patient-grouped out-of-fold discrimination, calibration, and fixed-threshold metrics. We further examined whether period-specific handcrafted features and sliding-window modeling emphasized similar overnight periods and physiological signals. Source-stratified and higher-burden endpoint analyses examined whether performance patterns varied by sensor source and outcome definition.

\begin{figure*}[ht]
  \centering
  \includegraphics[width=0.78\textwidth]{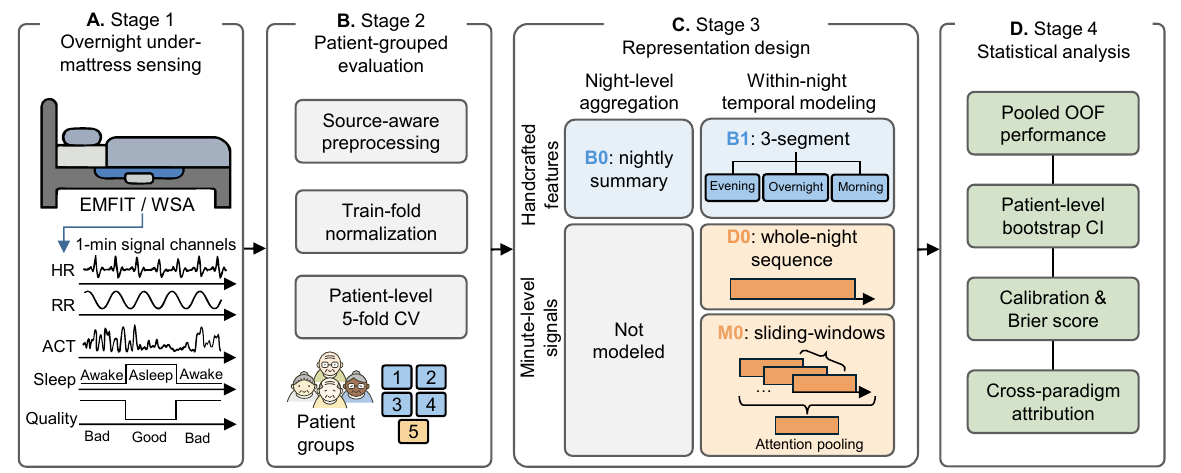}
  \caption{Four-paradigm modeling pipeline for retrospective next-day agitation risk scoring. The design contrasts handcrafted features with minute-level signals and whole-night aggregation with within-night representations. B0 uses nightly handcrafted summaries. B1 uses handcrafted features from the evening period (17:00--22:00), core overnight period (22:00--06:00), and morning period (06:00--10:00). D0 encodes the full ordered minute-level sequence. M0 encodes overlapping minute-level windows followed by attention pooling. All paradigms use the same patient-grouped folds and are compared using pooled OOF predictions.}
  \label{fig:workflow}
\end{figure*}

\section{Data and Methods}
\label{sec:methods}

Figure~\ref{fig:workflow} summarizes the study workflow, from under-mattress sensing and night-to-day label pairing to patient-grouped evaluation, pooled out-of-fold (OOF) analysis, calibration assessment, and cross-paradigm attribution.

\subsection{Study Cohort and Outcome}
\label{subsec:cohort}
This study analyzed 423 patient-nights from 65 patients, pairing overnight recordings from two passive under-mattress sensing systems (EMFIT and WSA) with next-day agitation assessments. Under the hospital monitoring protocol described by Davidoff et al.~\cite{davidoff2022toward}, continued sensor monitoring and agitation-label collection expanded the interim dataset analyzed by Liu et al.~\cite{liu2026dissociating} (55 participants and 333 patient-nights) to the current cohort. The study protocol was approved by the Ethics Committee Research of University Hospitals Leuven (reference ID S62882). Written informed consent was obtained from participants with capacity or, when cognitive impairment precluded consent, from a legally authorized representative, with study information explained to participants in a manner appropriate to their cognitive abilities.

The two sensor systems provided heart rate (HR), respiratory rate (RR), bed activity (ACT), and sleep-stage streams at different temporal resolutions. The preprocessing pipeline harmonized these streams to 1-minute resolution and derived a minute-level quality channel from device-specific signal-availability and missingness patterns. Supplementary Section S1 describes the device-specific cleaning of the 1-minute sequences, including overnight-record construction, quality harmonization, missingness marking, and source-aware normalization. Table~\ref{tab:cohort} summarizes the analysis cohort.

\begin{table}[ht]
\centering
\caption{Cohort characteristics and sensing configuration for the pooled EMFIT and WSA in-hospital analysis.}
\label{tab:cohort}
\small
\begin{tabular}{@{}p{0.42\columnwidth}p{0.53\columnwidth}@{}}
\toprule
Characteristic & Value \\
\midrule
\multicolumn{2}{@{}l}{\textit{Cohort and outcome}} \\
Patients & 65 \\
Patient-nights & 423 \\
Agitation-positive nights & 306 (72.3\%) \\
\addlinespace[2pt]
\hline
\multicolumn{2}{@{}l}{\textit{Demographic and cognitive characteristics}} \\
Male & 41 (63.1\%) \\
Female & 24 (36.9\%) \\
Age, years & $80.0 \pm 8.3$ (57--96) \\
MMSE score & $11.6 \pm 6.7$ (0--27; $n=48$) \\
\addlinespace[2pt]
\hline
\multicolumn{2}{@{}l}{\textit{Dementia diagnosis}} \\
AD & 34 (52.3\%) \\
NOS & 11 (16.9\%) \\
DLB & 6 (9.2\%) \\
VaD & 6 (9.2\%) \\
Mixed & 4 (6.2\%) \\
PDD & 2 (3.1\%) \\
Other & 2 (3.1\%) \\
\addlinespace[2pt]
\hline
\multicolumn{2}{@{}l}{\textit{Monitoring setting and sensing configuration}} \\
Setting & Specialized hospital dementia unit \\
Sensor placement & Under-mattress sensing \\
EMFIT patients / nights & 40 / 214 \\
WSA patients / nights & 25 / 209 \\
Temporal resolution & 1 min \\
Channels & HR, RR, ACT, sleep stage, quality \\
\bottomrule
\multicolumn{2}{@{}p{0.95\columnwidth}@{}}{\footnotesize \textit{Note:} Values are mean $\pm$ standard deviation (range) or $n$ (\%). MMSE was available for 48 patients (17 missing). AD = Alzheimer's disease; DLB = dementia with Lewy bodies; MMSE = Mini-Mental State Examination; NOS = not otherwise specified; PDD = Parkinson's disease dementia; VaD = vascular dementia; Mixed = Alzheimer's + vascular dementia; HR = heart rate; RR=respiratory  rate; ACT = bed activity; WSA = Withings Sleep Analyzer.} \\
\end{tabular}
\end{table}

Agitation was scored by nursing and research staff using the Pittsburgh Agitation Scale (PAS)~\cite{rosen1994pas} within the previously described signal- and event-contingent ecological momentary assessment protocol~\cite{davidoff2022toward,davidoff2024physiological}. Each overnight recording on night \(j\) was paired with agitation observations from the following daytime study period on day \(j+1\). Full participant monitoring, annotation, and data-collection procedures are reported by Davidoff et al.~\cite{davidoff2022toward}. PAS rates motor agitation, verbal agitation, aggression, and resistance to care as four domains scored from 0 to 4. The day-level PAS total score was calculated as the sum of the maximum daytime score in each domain. The primary endpoint was intentionally sensitive: a patient-day was classified as positive if any PAS domain score exceeded zero. It therefore includes minor or transient behaviors that may not prompt a change in care and should be interpreted as any recorded agitation rather than clinically significant agitation.
The resulting prevalence was 72.3\% (306/423 nights), reflecting a specialized hospital dementia-unit cohort with substantial behavioral and psychological symptoms of dementia (BPSD) and intensive behavioral monitoring rather than a general community-dwelling dementia population~\cite{cerejeira2012bpsd}.

\subsection{Four-Paradigm Modeling Pipeline}
\label{subsec:paradigms}
The four paradigms compare how alternative representations of one overnight sensor record are mapped to a next-day agitation probability (Fig.~\ref{fig:workflow}). B0 and B1 use handcrafted features at different levels of temporal aggregation, whereas D0 and M0 learn representations from minute-level signals.

To formalize this comparison, let \(\mathcal O_i\) be patient-night \(i\), \(\mathbf X_i\in\mathbb R^{T_i\times 5}\) its cleaned sequence of HR, RR, ACT, binary sleep--wake, and quality, and \(y_i\in\{0,1\}\) its following-day agitation outcome. Paradigm \(m\) estimates
\begin{equation}
\hat p_i^{(m)}=f_m\!\left(\mathcal R_m(\mathcal O_i)\right),
\quad m\in\{\mathrm{B0},\mathrm{B1},\mathrm{D0},\mathrm{M0}\},
\label{eq:common_prediction}
\end{equation}
where \(\mathcal R_m\) is the paradigm-specific representation and \(f_m\) is its selected prediction rule, including fold-local preprocessing and feature selection where applicable. The four representations can be summarized as
\begin{equation}
\begin{aligned}
\mathcal R_{\mathrm{B0}}(\mathcal O_i)
  &=\Phi(\mathcal O_i),\\
\mathcal R_{\mathrm{B1}}(\mathcal O_i)
  &=[\Phi(\mathcal O_i^{(E)}),\Phi(\mathcal O_i^{(C)}),\Phi(\mathcal O_i^{(M)}),\boldsymbol\Delta_i],\\
\mathcal R_{\mathrm{D0}}(\mathcal O_i)
  &=\mathbf X_i,\\
\mathcal R_{\mathrm{M0}}(\mathcal O_i)
  &=\sum_{j=1}^{J_i}a_{ij}\mathbf h_{ij},
  \quad \mathbf h_{ij}=e_\theta(\mathbf W_{ij}).
\end{aligned}
\label{eq:four_representations}
\end{equation}
Here, \(\Phi\) denotes handcrafted feature extraction; \(E\), \(C\), and \(M\) denote the evening, core overnight, and morning periods; and \(\boldsymbol\Delta_i\) contains cross-period descriptors. For M0, \(\mathbf W_{ij}\) is the \(j\)th valid 90-minute window obtained with a 45-minute stride, \(e_\theta\) is the window encoder, and \(J_i\) is the number of valid windows. The learned gated-attention score \(s_\psi\) produces normalized weights
\begin{equation}
a_{ij}
  =\frac{\exp\{s_\psi(\mathbf h_{ij})\}}
  {\sum_{k=1}^{J_i}\exp\{s_\psi(\mathbf h_{ik})\}}.
\label{eq:m0_attention}
\end{equation}
The weights satisfy \(\sum_{j=1}^{J_i}a_{ij}=1\) and quantify relative model-associated contribution to the pooled night representation. They should not be interpreted as causal or physiological effect estimates. A separate minute-level-input/night-level-summary pipeline was not evaluated because aggregating the minute-level streams over the full night would substantially overlap with B0.

\textbf{B0 -- Nightly summary} (\emph{handcrafted features $\times$ whole-night aggregation}). The operator \(\Phi\) summarizes HR, RR, sleep architecture, activity, and signal quality over the full night~\cite{liu2026dissociating}. Within each training fold, SHAP-based feature selection (SHAP-FS) ranks features by their mean absolute SHAP value for the positive class, reflecting their average attribution magnitude within the selector model~\cite{lundberg2017shap}. The selected representative is a Random Forest using fold-wise SHAP-FS.

\textbf{B1 -- Period-resolved summary} (\emph{handcrafted features $\times$ period-resolved aggregation}). B1 applies \(\Phi\) separately to the evening period (17:00--22:00), core overnight period (22:00--06:00), and morning period (06:00--10:00), then appends \(\boldsymbol\Delta_i\) to capture cross-period changes. Using the same fold-wise SHAP-FS procedure as B0, the selected representative is Logistic Regression. To standardize tabular input dimensionality and limit feature-space complexity, B0 and B1 use the top 30 fold-specific SHAP-ranked features. Supplementary Section S2 details the period definitions and cross-period descriptors.

\textbf{D0 -- Whole-night raw-sequence modeling} (\emph{minute-level signals $\times$ full-sequence modeling}). D0 directly models \(\mathbf X_i\) as the complete ordered minute-level sequence. The selected representative is InceptionTime~\cite{ismail2020inceptiontime}.

\textbf{M0 -- Sliding-window MIL} (\emph{minute-level signals $\times$ window-based MIL}). M0 partitions \(\mathbf X_i\) into overlapping 90-minute windows with a 45-minute stride. The selected representative uses a Transformer window encoder with gated-attention MIL pooling~\cite{phan2022sleeptransformer,ilse2018attention}. Supplementary Section S6 reports the window/stride sensitivity analysis.

Candidate model families were screened according to held-out validation performance using the predefined patient-grouped folds, and one representative model was selected for each pipeline. For B0 and B1, selection was performed within the prespecified SHAP-FS configuration. M0 window and stride were selected using the same mean-fold criterion. Further implementation and candidate-model screening details are provided in Supplementary Sections S3--S6. Final performance is summarized from pooled OOF predictions.

\subsection{Preprocessing and Cross-Validation}
\label{subsec:preprocessing}
Each overnight record comprised four 1-minute data streams: heart rate, respiratory rate, bed activity, and a binary sleep--wake channel derived from the device-provided Deep, Light, REM, and Wake/Awake stages. Deep, Light, and rapid eye movement (REM) were encoded as sleep (1), whereas Wake/Awake was encoded as wake (0); missing or unrecognized stages were also encoded as 0, consistent with the implemented preprocessing. A fifth quality channel was derived from device-specific signal-availability and missingness rules. B0 and B1 used statistical and sleep-related features extracted over either the full night or three predefined overnight periods, whereas D0 and M0 received the cleaned 1-minute channels directly. EMFIT and WSA differ in activity scale, derived sleep metrics, and signal dropout characteristics~\cite{ravindran2024reliable,liu2026dissociating}. Derived tabular features and continuous sequence channels were therefore imputed and normalized separately by source using parameters estimated only from each training fold. The binary sleep--wake and quality sequence channels remained on their original scales. Detailed cleaning, quality construction, feature extraction, and normalization procedures are provided in Supplementary Section S1.

All four pipelines used the same predefined five-fold patient-grouped assignment, constructed to balance sensor source and the distribution of patient-level agitation rates across folds. All nights from a patient remained within one fold, and each patient-night contributed exactly one held-out OOF prediction. Class-imbalance weights were estimated exclusively from the corresponding training fold, with no imbalance parameter tuned on held-out predictions.

\subsection{Performance Evaluation}

The evaluation was designed to distinguish ranking performance, probability error, and fixed-threshold operating behavior, which can diverge for a high-prevalence outcome (72.3\% agitation-positive patient-nights in this cohort). Performance was evaluated from pooled OOF predictions using area under the receiver operating characteristic curve (AUROC), area under the precision--recall curve (AUPRC), Brier score, sensitivity, specificity, and balanced accuracy at a fixed probability threshold of 0.5. Balanced accuracy is the mean of sensitivity and specificity. Brier score summarizes probability error, and calibration is visualized with reliability diagrams using 10 quantile-based probability bins.

Given this event prevalence, three reference baselines were included to show how the prevalence of agitation-positive nights affects probability-based and fixed-threshold metrics. For a held-out fold with training-fold prevalence $\pi_{\mathrm{train}}$, the always-positive baseline assigned probability 1, the train-prevalence baseline assigned the constant probability $\pi_{\mathrm{train}}$, and the stratified-random baseline sampled predictions from $\mathrm{Bernoulli}(\pi_{\mathrm{train}})$.

\subsection{Statistical Analysis}
\label{subsec:stats}
Statistical analysis was designed to quantify uncertainty in model performance and paired between-pipeline differences while preserving the dependence among repeated nights from the same patient. Confidence intervals were estimated from 1000 patient-level bootstrap replicates, and paired model differences were computed within the same bootstrap samples. Paired DeLong tests~\cite{delong1988comparing} were additionally reported as unadjusted secondary AUROC comparisons for B1--B0, D0--B0, M0--B0, D0--B1, M0--B1, and M0--D0. Patient-level bootstrap intervals remained the primary uncertainty estimates because they account for clustering by patient.

\subsection{Cross-Model Feature Attribution Analysis}
\label{subsec:attribution_methods}

We compared B1 and M0 to assess whether two contrasting within-night representations assigned relatively high model-associated importance to similar overnight periods and signals. The comparison focused on these pipelines because their importance scores could be summarized over the same three periods and five signals. The resulting analysis comprised 15 period--signal combinations defined by the evening, core overnight, and morning periods and by HR, RR, ACT, the binary sleep--wake channel, and quality. For B1, absolute fold-specific SHAP values for segment-derived features corresponding to a single period and signal were grouped and averaged across OOF nights. For M0, prediction sensitivity was measured by replacing one signal within one period with its preprocessing reference value and calculating the absolute change in predicted probability. Since SHAP magnitude and perturbation-based probability change are measured on different scales, the 15 combinations were converted to within-model relative ranks and displayed in a rank--rank scatter plot. Spearman correlation summarized descriptive agreement between the rankings. This descriptive rank comparison is hypothesis-generating and does not establish physiological mechanisms or causal relationships.

\subsection{Subgroup and Sensitivity Analyses}
Since the pooled cohort incorporated recordings from two sleep sensors and the primary endpoint used a broad definition of agitation, these additional analyses examined performance variation across sensor sources and under a higher-burden agitation definition.

Source-stratified performance was assessed by calculating metrics separately for EMFIT and WSA from the pooled patient-grouped OOF predictions. Both sensor subsets were represented within the common patient-grouped cross-validation framework.

The primary endpoint classified a night as positive when any next-day PAS domain score exceeded zero. Endpoint sensitivity was examined using a day-level PAS total score greater than 4. Because an individual PAS domain has a maximum score of 4, a day-level PAS total score greater than 4 requires agitation in at least two domains and at least one domain score above 1. This endpoint therefore represents more extensive or more intense agitation than the primary endpoint. The sensitivity analysis tests performance under a lower-prevalence, higher-burden outcome.

\section{Results}
\label{sec:results}

\subsection{Overall Predictive Performance}
\label{subsec:main_results}
The pooled OOF analysis separated the four prespecified pipelines into two broad performance patterns (Table~\ref{tab:main_results}). D0 and M0 had higher discrimination point estimates and more balanced fixed-threshold behavior than B0 and B1. D0 achieved the highest AUROC (0.692 [0.622, 0.759]), AUPRC, specificity, and balanced accuracy, while M0 achieved similar balanced accuracy. B1 had higher AUROC and specificity point estimates than B0.

\begin{table*}[!htb]
\centering
\caption{Main pooled OOF results for the EMFIT and WSA in-hospital cohort (423 nights, prevalence 0.723). Metrics are reported as pooled OOF point estimates with 95\% CIs from patient-level bootstrap resampling. Boldface indicates the best performance among B0/B1/D0/M0.}
\label{tab:main_results}
\scriptsize
\setlength{\tabcolsep}{2.0pt}
\resizebox{\textwidth}{!}{%
\begin{tabular}{lcccccc}
\toprule
Model & AUROC & AUPRC & Sensitivity & Specificity & Brier & BalAcc \\
\midrule
Always positive & 0.500 [0.500, 0.500] & 0.723 [0.648, 0.796] & 1.000 [1.000, 1.000] & 0.000 [0.000, 0.000] & 0.277 [0.204, 0.352] & 0.500 [0.500, 0.500] \\
Train prevalence & 0.441 [0.333, 0.554] & 0.693 [0.594, 0.795] & 1.000 [1.000, 1.000] & 0.000 [0.000, 0.000] & 0.201 [0.169, 0.235] & 0.500 [0.500, 0.500] \\
Stratified random & 0.516 [0.463, 0.572] & 0.730 [0.652, 0.806] & 0.732 [0.681, 0.787] & 0.299 [0.211, 0.400] & 0.388 [0.334, 0.447] & 0.516 [0.463, 0.572] \\
B0-RF & 0.565 [0.472, 0.657] & 0.785 [0.690, 0.866] & \textbf{0.948 [0.907, 0.980]} & 0.077 [0.031, 0.126] & \textbf{0.206 [0.174, 0.240]} & 0.512 [0.482, 0.543] \\
B1-LR & 0.622 [0.547, 0.703] & 0.816 [0.737, 0.881] & 0.644 [0.569, 0.727] & 0.538 [0.414, 0.658] & 0.246 [0.212, 0.278] & 0.591 [0.525, 0.662] \\
D0-InceptionTime & \textbf{0.692 [0.622, 0.759]} & \textbf{0.849 [0.778, 0.907]} & 0.641 [0.541, 0.745] & \textbf{0.675 [0.562, 0.777]} & 0.218 [0.193, 0.241] & \textbf{0.658 [0.598, 0.718]} \\
M0-Transformer & 0.679 [0.592, 0.764] & 0.812 [0.733, 0.889] & 0.680 [0.582, 0.775] & 0.632 [0.490, 0.770] & 0.223 [0.197, 0.249] & 0.656 [0.585, 0.720] \\
\bottomrule
\end{tabular}
}
\end{table*}

\subsection{Pairwise Statistical Comparisons}
\label{subsec:paired_results}
Paired patient-level bootstrap comparisons localized the clearest discrimination gains to the sequence- and window-based models relative to B0: the AUROC differences were +0.127 for D0 and +0.114 for M0, with both confidence intervals excluding zero (Table~\ref{tab:paired_comparisons}). The remaining paired intervals crossed zero, including comparisons of D0 and M0 with B1 and with each other. DeLong tests are reported as secondary comparisons.

\begin{table}[t]
\centering
\caption{Paired AUROC comparisons on shared pooled OOF predictions. Patient-level bootstrap intervals are the primary uncertainty estimates.}
\label{tab:paired_comparisons}
\footnotesize
\begin{tabular}{lccc}
\toprule
Comparison & $\Delta$AUROC & Bootstrap 95\% CI & DeLong $p$ \\
\midrule
B1--B0 & +0.057 & [-0.019, +0.133] & 0.077 \\
D0--B0 & +0.127 & [+0.037, +0.219] & 0.0005 \\
M0--B0 & +0.114 & [+0.021, +0.210] & 0.0015 \\
D0--B1 & +0.070 & [-0.022, +0.163] & 0.057 \\
M0--B1 & +0.057 & [-0.037, +0.144] & 0.113 \\
M0--D0 & -0.013 & [-0.103, +0.077] & 0.714 \\
\bottomrule
\end{tabular}
\end{table}

\subsection{Calibration and Prevalence-based Baselines}

\begin{figure*}[t]
  \centering
  \begin{minipage}[t]{0.62\textwidth}
    \centering
    \includegraphics[width=\linewidth]{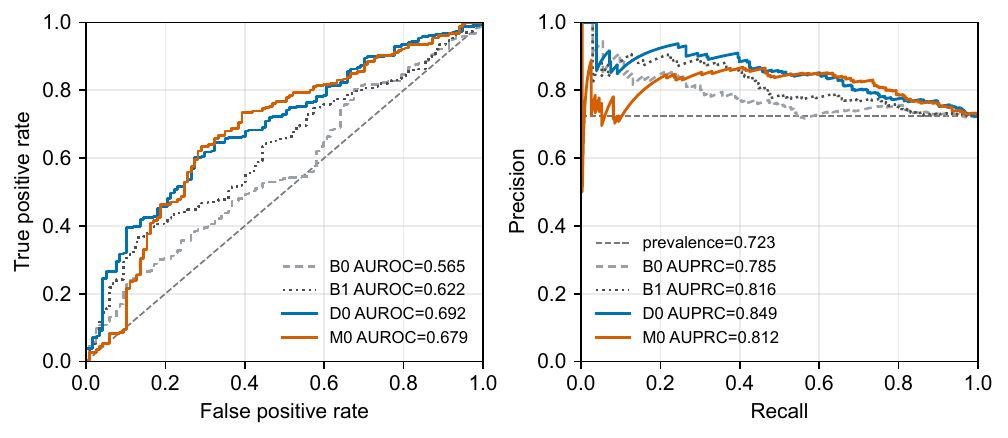}
    \vspace{-0.5em}
    {\small (a) ROC and precision-recall overlay.}
    \label{fig:roc_pr}
  \end{minipage}\hfill
  \begin{minipage}[t]{0.31\textwidth}
    \centering
    \includegraphics[width=0.93\linewidth]{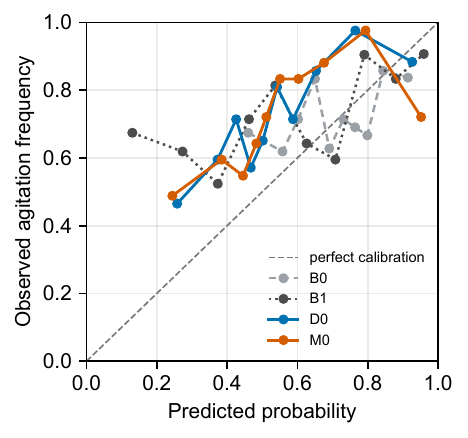}
    \vspace{-0.5em}
    {\small (b) Reliability diagram.}
    \label{fig:reliability}
  \end{minipage}
  \caption{Discrimination and calibration on shared pooled OOF predictions. D0 has the highest AUROC, AUPRC, and balanced accuracy, whereas B0 has the lowest learned-model Brier score despite weak specificity. Calibration remains limited relative to the train-prevalence probability baseline in Table~\ref{tab:main_results}.}
  \label{fig:roc_calibration}
\end{figure*}

The simple reference baselines contextualize the impact of the high prevalence of agitation-positive nights (72.3\%). The always-positive and train-prevalence baselines had no useful discrimination or specificity, although the train-prevalence baseline achieved the lowest Brier score overall because its constant probability closely matched the pooled event rate (Table~\ref{tab:main_results}). B0 produced a similar Brier score but classified most nights as positive, yielding high sensitivity and very low specificity. Its relatively low probability error therefore did not yield balanced classification at the prespecified 0.5 threshold. The reliability diagram indicates that calibration remained limited across the learned models (Fig.~\ref{fig:roc_calibration}).

\subsection{Cross-Model Feature Attribution}
\label{subsec:mechanism}
Feature attribution from the B1 Logistic Regression and M0 MIL Transformer was summarized across 15 overnight-period--signal combinations and converted to within-model relative ranks for qualitative comparison.

\begin{figure}[t]
  \centering
  \includegraphics[width=0.7\linewidth]{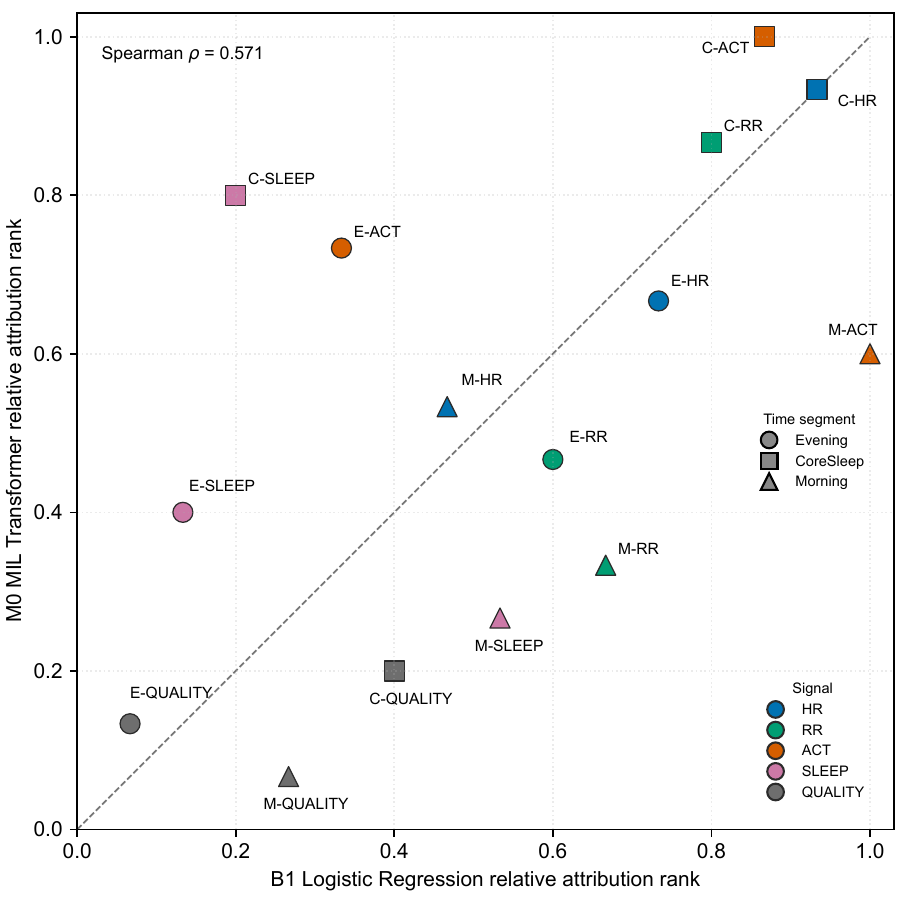}
  \caption{Qualitative comparison of B1 segment-feature and M0 final-model attribution rankings across 15 segment--signal combinations. Each axis shows the relative attribution rank within one model, with higher values indicating higher rank. Combinations near the dashed diagonal have similar ranks, and those in the upper-right region rank highly in both models. Marker shape denotes the within-night segment and color denotes the signal. The descriptive Spearman rank correlation was $\rho=0.57$.}
  \label{fig:consistency}
\end{figure}

The final-model attribution summaries showed moderate descriptive rank agreement (Spearman $\rho=0.57$, Fig.~\ref{fig:consistency}). ACT, HR, and RR during the core overnight period occupied the upper-right region, indicating relatively high attribution ranks in both models. Morning-period ACT and the binary sleep--wake channel during the core overnight period lay farther from the diagonal. Morning-period ACT ranked higher in B1, whereas the core-overnight sleep--wake channel ranked higher in M0.

\subsection{Source-Stratified Performance}
\label{subsec:boundary_results}
Source-stratified performance varied between EMFIT and WSA (Table~\ref{tab:source_stratified}). B0 showed the largest source-stratified AUROC difference, while M0 also differed across the two subsets. B1 and D0 were comparatively stable in AUROC. Despite stable discrimination, D0's balanced accuracy was lower for WSA, primarily because of lower specificity.

\begin{table}[t]
\centering
\caption{Sensor-source-stratified performance from pooled OOF predictions. EMFIT included 214 nights with a prevalence of 0.673, and WSA included 209 nights with a prevalence of 0.775.}
\label{tab:source_stratified}
\footnotesize
\setlength{\tabcolsep}{1.8pt}
\begin{tabular}{@{}llccccc@{}}
\toprule
Source & Model & AUROC & AUPRC & Sens. & Spec. & BalAcc \\
\midrule
\multirow{4}{*}{EMFIT} & B0 & 0.506 & 0.679 & 0.938 & 0.057 & 0.497 \\
& B1 & 0.606 & 0.749 & 0.611 & 0.557 & 0.584 \\
& D0 & 0.688 & 0.802 & 0.625 & 0.743 & 0.684 \\
& M0 & 0.648 & 0.758 & 0.597 & 0.643 & 0.620 \\
\midrule
\multirow{4}{*}{WSA} & B0 & 0.631 & 0.866 & 0.957 & 0.106 & 0.532 \\
& B1 & 0.630 & 0.868 & 0.673 & 0.511 & 0.592 \\
& D0 & 0.680 & 0.881 & 0.654 & 0.574 & 0.614 \\
& M0 & 0.695 & 0.851 & 0.753 & 0.617 & 0.685 \\
\bottomrule
\end{tabular}
\end{table}

\subsection{Sensitivity to Agitation Endpoint Definition}

Defining higher-burden agitation as a day-level PAS total score greater than 4 reduced outcome prevalence from 72.3\% to 17.3\% (Supplementary Table S3). D0 retained the highest AUROC (0.654), AUPRC (0.286), sensitivity (0.575), and balanced accuracy (0.638), whereas B1 achieved the lowest Brier score (0.164) and highest specificity (0.911). The analysis therefore showed partial stability in model ranking but substantial changes in operating characteristics under the lower-prevalence endpoint.

\section{Discussion}
\label{sec:discussion}
\subsection{Temporal Representation and Model Behavior}
The principal finding is that contactless under-mattress signals from the preceding night contained modest information for estimating next-day agitation risk among PLWD in a hospital setting. Representations retaining minute-level within-night structure achieved better discrimination than conventional whole-night summaries, and performance varied according to how that temporal structure was represented. Their AUROC gains over B0 were supported by patient-level bootstrap intervals, whereas differences relative to the period-specific B1 pipeline and between D0 and M0 remained uncertain. The overall predictive performance was modest, and the comparison does not isolate representation effects from classifier--representation interactions.

D0 and M0 retained temporal variation at different scales: D0 modeled the complete ordered minute-level sequence and achieved the highest AUROC, AUPRC, specificity, and balanced accuracy among the four evaluated pipelines, whereas M0 encoded local windows before learned aggregation and remained close to D0 in balanced accuracy. In contrast, B0 applies the greatest temporal compression, while B1 retains only a coarse predefined partition of the night. The four-pipeline benchmark therefore characterizes the consequences of different levels of temporal compression under a common patient-grouped design and identifies full-sequence and window-based approaches as stronger candidates than nightly summaries in this cohort. The data do not, however, establish that either learned temporal strategy is definitively superior.

This representation-centered comparison extends prior work in which LightGBM was selected for engineered daily or multi-day feature sets and complements the earlier mixed-effects analysis of feature-level associations~\cite{ramesh2025,bafaloukou2025interpretable,liu2026dissociating}. Because the analytical tasks, input representations, and evaluation designs differed, performance values from the predictive studies should not be compared directly with those from the present patient-grouped benchmark.

The prevalence-based baselines demonstrate how the high event rate (72.3\%) influences predictive metrics. The always-positive and train-prevalence baselines had zero specificity, while the train-prevalence baseline achieved the lowest Brier score despite having no useful discrimination. B0 showed a related pattern, combining high sensitivity with very low specificity and a Brier score close to the prevalence baseline. These findings demonstrate that sensitivity or probability error alone may overstate performance in a high-prevalence cohort and underscore the more balanced discrimination of D0 and M0. Calibration nevertheless remained limited across the learned models.

\subsection{Physiological Context of Nocturnal Attribution Patterns}
The attribution ranking comparison identified activity, heart rate, and respiratory rate during the core overnight period as common high-ranking period--signal combinations across B1 and M0. Their convergence across two different representations prioritizes these signals for targeted physiological validation beyond static nightly summaries. Movement-derived measures during this period may reflect sleep fragmentation or nocturnal restlessness~\cite{lim2013sleep,saleh2025actigraphy}, both of which have been linked to daytime behavioral symptoms in PLWD~\cite{brown2015sleep}.

The high attribution ranks of HR and respiratory rate during the core overnight period suggest that nocturnal cardiorespiratory variation may contain information relevant to next-day agitation. This interpretation is consistent with related contactless-sensing evidence associating nocturnal respiratory measures with next-day agitation occurrence~\cite{liu2026dissociating}, and with broader in-home monitoring work showing that sleep and physiological features can contribute to agitation prediction~\cite{bafaloukou2025interpretable}. Previous work linking heart rate variability with agitation risk in Alzheimer's disease provides complementary evidence that autonomic physiology may be relevant, although at a substantially longer timescale~\cite{liu2023ANSAgit}. These descriptive attribution patterns generate hypotheses about sleep disruption, autonomic physiology, and other potential mechanisms that require further validation.

The largest rank differences provide complementary insight into how temporal representation shapes model behavior. Activity during the morning period ranked higher in B1, suggesting that segment-level summary features may emphasize concentrated activity around the transition out of the overnight period. In contrast, the binary sleep--wake channel during the core overnight period ranked higher in M0, suggesting greater sensitivity to local sleep--wake composition than aggregated segment features capture. These representation-specific differences complement the shared emphasis on activity and cardiorespiratory signals during the core overnight period.

\subsection{Robustness and Generalizability}
Patient-grouped evaluation tests whether signals from the preceding night can support risk scoring for patients excluded from the corresponding training fold. Accordingly, the evidence pertains to cross-patient risk estimation and does not establish patient-specific day-to-day forecasting.

Source-stratified performance point estimates differed across EMFIT and WSA. Since the data were collected in the same ward during different periods, observed variation may reflect differences in device hardware, device-specific signal processing, labeling protocols, and patient cohorts. D0 rank discrimination was comparatively stable across the two acquisition streams, but the analysis does not establish transportability to other devices or clinical settings.

Model applicability also depends on how agitation is defined. The higher-burden endpoint preserved D0's leading rank on several metrics but changed prevalence and operating characteristics substantially. These findings show that model behavior and practical interpretation depend on the selected agitation definition.

\subsection{Clinical Interpretation}
The primary endpoint was intentionally sensitive and classified any nonzero PAS domain score as positive. It therefore includes minor or transient behaviors that may not prompt a change in care and should be interpreted as any recorded agitation rather than clinically significant agitation. A positive model output consequently represents an elevated probability of any observed agitated behavior under the study definition, not a diagnosis of clinically actionable agitation. The changes in prevalence and operating characteristics under the higher-burden endpoint show that performance depends on the intended behavioral target as well as the sensing representation.

Agitation in PLWD can reflect multiple and potentially modifiable contributors, including unmet needs, discomfort, and clinical or environmental stressors~\cite{kales2015agit_assess,cohen2015unmet}. A future sensor-based alert should therefore be treated as an adjunct to observation and structured assessment rather than as a stand-alone indication for treatment. Its most defensible role would be to prompt timely review of the patient and care context when staff judgment also indicates concern.

Limited calibration means that the predicted probabilities cannot yet be interpreted as reliable absolute risks; for example, a predicted risk of 70\% does not necessarily correspond to a 70\% observed probability of next-day agitation~\cite{vanCalster2019calibration}. At the present performance level, a high-risk score should not trigger treatment. Before workflow use, calibration and the operating threshold would need prospective evaluation in the target setting because the consequences of missed events and unnecessary alerts depend on the intended response.

\subsection{Strengths, Limitations, and Future Work}
The study combines 423 repeated overnight observations acquired during routine hospital care, two passive under-mattress sensing systems, minute-level physiological and activity signals, and a common patient-grouped OOF evaluation across four temporal representations. Source-aware preprocessing and patient-level bootstrap resampling address device heterogeneity and within-patient dependence. Together, these elements provide a dense short-term repeated-measures dataset under real-world acquisition conditions.

The study contains a substantial number of repeated overnight observations, but cross-patient generalizability remains limited by the 65 independent participants and recruitment from a single specialized hospital dementia unit. Inter-rater reliability was not assessed, potentially introducing variability into the outcome labels due to differences between nursing and research staff ratings. The broad primary endpoint also captures minor or transient behaviors, limiting inference about clinically significant agitation. In addition, the representative classifiers and M0 windowing configuration were selected through exploratory comparisons on the same patient-grouped folds rather than through a separate model-selection cohort or nested cross-validation. Consequently, reported performance may be optimistic in future samples, and the bootstrap confidence intervals do not include uncertainty introduced by configuration selection.

Future work should use prespecified model selection, external validation across hospital and device settings, prospective and source-stratified calibration, and targeted physiological validation of the period--signal combinations that ranked highly in our cross-model attribution analysis. These steps are required to establish robustness beyond this single-site cohort and to determine whether the risk scores can support a defined care workflow.

\section{Conclusion}
\label{sec:conclusion}
Under-mattress signals recorded during the preceding night contained modest information about next-day agitation risk in this in-hospital dementia cohort. Full-sequence and sliding-window modeling achieved stronger discrimination than conventional nightly summaries, although their advantages over period-specific engineered features remained uncertain. Cross-model attribution consistently prioritized activity and cardiorespiratory signals during the core overnight period for further physiological validation. These findings support minute-level temporal modeling as a promising engineering direction, but external validation and prospective calibration are required before the resulting scores can guide individual care decisions.

\section*{Acknowledgments}
The authors thank all participating patients and their families, as well as the nursing staff and psychiatrists in training at Cog K of UPC KU Leuven.

\section*{Statements and declarations}
\subsection*{Ethical considerations}
Ethical approval for this study was granted by the Ethics Committee Research of University Hospitals Leuven (reference ID S62882). Given the participants' diagnosis of dementia, written informed consent was primarily secured through legal representatives. However, to ensure patient inclusion, the study details were explained verbally in a manner commensurate with each patient’s cognitive abilities. All patient data were pseudonymized and handled with strict confidentiality. Participation was voluntary and uncompensated, with the explicit understanding that the decision to enroll would not impact the standard of care received during admission.

\subsection*{Consent to participate} 
Informed consent was obtained from all individual participants included in the study. For participants who were unable to provide consent due to cognitive impairment, written
informed consent was obtained from their legal representativesor primary caregivers. All procedures were performed in accordance with the study protocol approved by the Research Ethics Committee of University Hospitals Leuven.

\subsection*{Funding statement}
This work was supported by the Flanders AI Research (FAIR) Program; the Research Foundation – Flanders (FWO) under grant number G046925N; the `Funds Malou Malou, Perano, Georgette Paulus, JMJS Breugelmans and Gabrielle, François and Christian De Mesmaeker' managed by the King Baudouin Foundation of Belgium (2021-J1990130-222081); KU Leuven (grants C2M/23/053, CELSA/24/019, and the Global PhD partnerships with the University of Melbourne GPUM/22/022); the Paris Brain Institute and KU Leuven (Big Brain Theory 4 grant); the `Klinische onderzoeks- en opleidingsraad (KOOR)' of the University Hospitals Leuven; and the AI-PROGNOSIS project under the European Union’s Horizon Europe research and innovation program (Grant Agreement No. 101080581). Views and opinions expressed are however those of the author(s) only and do not necessarily reflect those of the European Union. Neither the European Union nor the granting authority can be held responsible for them.

\subsection*{Data and code availability}
The primary dataset analyzed in the current study is not publicly available due to patient privacy and ethical restrictions, but de-identified data are available from the corresponding author upon reasonable request.

The code for experiments presented in this study will be made available by the corresponding author upon reasonable request.

\subsection*{Declaration of generative AI use}
During the preparation of this work, the authors used OpenAI's ChatGPT for English editing, code debugging, and optimization during manuscript preparation. All content was reviewed and verified by the authors, who retain full responsibility for the final work.

\bibliographystyle{IEEEtran}
\bibliography{mylib}

\clearpage
\onecolumn

\setcounter{section}{0}
\setcounter{table}{0}
\setcounter{figure}{0}
\setcounter{equation}{0}
\renewcommand{\thesection}{S\arabic{section}}
\renewcommand{\thetable}{S\arabic{table}}
\renewcommand{\thefigure}{S\arabic{figure}}
\renewcommand{\theequation}{S\arabic{equation}}

\renewcommand{\theHsection}{supplement.section.\arabic{section}}
\renewcommand{\theHtable}{supplement.table.\arabic{table}}
\renewcommand{\theHfigure}{supplement.figure.\arabic{figure}}
\renewcommand{\theHequation}{supplement.equation.\arabic{equation}}

\begin{center}
{\LARGE\bfseries Supplementary Material}\par
\vspace{0.5em}
{\large Under-Mattress Temporal Sensing for Next-Day Agitation Risk Scoring in Dementia Wards}\par
\end{center}
\vspace{1em}

This document provides preprocessing and model-configuration details, together with secondary analyses that qualify the four-paradigm comparison in the main manuscript. It documents the four input representations and representative-model selection, followed by analyses of window sensitivity, the higher-burden agitation endpoint, patient-level heterogeneity, and illustrative single-subject trajectories.

\section{Temporal Sequence Preprocessing Details}
\label{app:temporal_preprocessing}

The temporal models are trained from cleaned 1-minute sequence files produced by device-specific raw-data processing. All device streams are restricted to an overnight window from 17:00 to the following 10:00 before modeling, and each resulting overnight record is paired with the subsequent daytime agitation label. Physiologically implausible HR and RR values are set unavailable, missingness indicators are included in the cleaned files, and severely unreliable minutes are marked by a 1-minute quality flag.

For EMFIT, raw vitals sampled at approximately 4-second resolution are range-checked using HR 35--180 bpm and RR 6--40 rpm, snapped to a complete 4-second grid, and aggregated to 1-minute resolution. HR and RR use median aggregation, while activity uses mean activity. The EMFIT quality flag is based on the 1-minute proportion of simultaneous HR/RR/ACT absence, with severe missingness marked when the all-signal missing proportion is at least 0.4 or the completed grid minute is empty. The EMFIT sleep-stage field is aligned from the sleep-class file to the vital stream and mapped to Deep, Light, REM, or Awake where vital data are present.

For WSA, the raw series is already at 1-minute resolution. HR and RR are range-checked with the same physiological bounds as EMFIT and then processed with a NaN-safe median filter with window size 3. The movement-score field is used as the activity channel and is not median-filtered. The WSA quality flag marks minutes in which HR, RR, and movement score are all missing, and the cleaned WSA state field provides the sleep-stage channel used by downstream feature extraction.

EMFIT and WSA are therefore treated as related but non-exchangeable measurement sources. Although HR and RR use the same physical units, the sensor streams differ in temporal resolution and device-specific processing. Activity is derived from device-specific activity or movement-score fields, sleep-stage fields originate from device-specific outputs, and the quality channel is constructed during preprocessing. Source-aware imputation and scaling are used to avoid forcing the resulting source-specific distributions onto a single pooled scale before model fitting.

The selected temporal channels are HR, RR, ACT, the binary sleep--wake channel, and quality. Device-provided Deep, Light, and REM stages are encoded as sleep (1), whereas Wake/Awake is encoded as wake (0). Missing or unrecognized stages are also encoded as 0, consistent with the implemented preprocessing. The quality channel is derived from the cleaned 1-minute severe-missingness flag. It is encoded as 0 for valid data and 1 for invalid or unknown data, and missing quality flags are treated conservatively as invalid. D0 loads each night as an ordered 1-minute sequence, center-cropped when longer than 800 minutes and right-padded when shorter. M0 splits each night into overlapping 90-minute windows with 45-minute stride before MIL pooling. For both D0 and M0, continuous channels are normalized within each training fold using source-specific robust median/MAD statistics from the training partition only, and held-out fold sequences are transformed with the corresponding training-fold statistics. Activity is winsorized at the within-night 1st and 99th percentiles, missing activity values are filled with zero before transformation, and \(\log(1+x)\) is applied. The sleep--wake and quality channels remain on their binary scales. D0 uses the validity mask during fold-specific normalization and represents padded timesteps as zeros during modeling, whereas M0 includes only windows meeting the minimum valid-data ratio (70\%) and uses the minute-level quality flag as an input channel.

\section{Three-Segment Handcrafted Feature Construction}
\label{app:three_segment_features}

B1 uses a fixed three-segment representation designed to preserve coarse within-night timing while remaining compatible with tabular classifiers. Each overnight record is partitioned into the evening period (17:00--22:00), core overnight period (22:00--06:00), and morning period (06:00--10:00). The corresponding feature-name prefixes are Evening, CoreSleep, and Morning. Within each period, handcrafted statistics are computed for HR, RR, ACT, the binary sleep--wake channel, and quality-derived channels. These include central tendency, dispersion, extrema, interquartile range, coefficient of variation, and distribution-shape summaries where defined, together with sleep-related and presence/unknown-ratio descriptors.

The B1 table also includes cross-period descriptors that encode directional or stability changes across the night. These features include HR dipping from the evening to the core overnight period, HR morning surge from the core overnight to the morning period, RR dipping from the evening to the core overnight period, ACT escalation from the evening to the morning period, and HR/RR/ACT variability ratios across periods. These descriptors are not treated as separate outcomes or mechanistic claims. They are tabular summaries of within-night temporal contrast. All B1 imputation, scaling, SHAP-based feature selection, and classifier fitting are performed within the training fold before held-out OOF prediction.

\section{B0/B1 Tabular Model Configurations}
\label{app:tabular_configs}

B0 and B1 use the same patient-grouped folds, source-aware preprocessing, candidate classifier family, top-30 SHAP-FS procedure, and mean held-out-fold AUROC selection criterion. B0 uses whole-night summary features, while B1 uses the three-segment feature table plus cross-segment descriptors. This shared screening protocol supports comparison of the complete B0 and B1 pipelines, but the different selected classifiers prevent attribution of performance differences to temporal representation alone. For both paradigms, imputation, scaling, feature selection, and classifier fitting are restricted to the training fold before held-out OOF probabilities are generated.

\textbf{Tuned models.} The tuned tabular analysis evaluates the same classifier set for B0 and B1: Logistic Regression, Random Forest, AdaBoost, SVM with RBF kernel, linear SVM, multilayer perceptron (MLP), XGBoost~\cite{chen2016xgboost}, and LightGBM~\cite{ke2017lightgbm}. Source-specific robust scaling is applied within each training fold. Logistic Regression uses class weighting and a maximum of 1000 iterations. Random Forest uses 100 trees, maximum depth 10, minimum split size 5, and balanced class weights. AdaBoost uses 100 depth-1 decision stumps and a learning rate of 0.1. The SVM models use probability output and balanced class weights. The MLP uses hidden layers $(64,32,16)$, ReLU activation, the Adam optimizer, $\alpha=0.001$, and a maximum of 500 iterations. XGBoost and LightGBM use 100 trees, maximum depth 6, learning rate 0.1, and fold-specific class-imbalance weighting. All imbalance parameters are estimated exclusively from training-fold labels. Mean held-out-fold AUROC is used to select the representative classifier within each prespecified tabular configuration. The primary input-dimension-standardized comparison selects B0 and B1 within the SHAP-FS configuration.

\textbf{SHAP-based feature selection.} Feature selection is performed independently within each training fold. A balanced Random Forest selector with 100 trees, maximum depth 10, and minimum split size 5 is trained on the fold-specific scaled training data, and features are ranked by mean absolute SHAP value for the positive class~\cite{lundberg2017shap}. To standardize tabular input dimensionality and limit feature-space complexity, both the B0 summary-feature representation and the B1 three-segment representation use the top 30 fold-specific SHAP-ranked features. The candidate classifier set and principal hyperparameters then match the tuned-model analysis. For gradient-boosted classifiers, the positive class is weighted by the training-fold ratio of negative to positive observations. This fold-local procedure prevents held-out-fold information from influencing feature selection. The primary B0 model is SHAP-selected Random Forest on nightly summary features, and the primary B1 model is SHAP-selected Logistic Regression on three-segment features.

\section{D0/M0 Temporal Model Configurations}
\label{app:temporal_configs}

Both D0 and M0 use the same five model-input channels: HR, RR, ACT, the binary sleep--wake channel, and the derived quality channel. They are evaluated using the same patient-grouped folds as B0 and B1. For each selected temporal representative, each held-out fold is evaluated once using the final model obtained after the fixed number of training epochs.

\textbf{D0 configuration.} D0 models each overnight record as one five-channel 1-minute sequence. Sequences are center-cropped or right-padded to 800 minutes, and fold-specific sequence normalization is fit on the training partition only. The candidate family includes ResNet1D~\cite{wang2017time}, InceptionTime~\cite{ismail2020inceptiontime,middlehurst2024bakeoff}, TCN~\cite{bai2018tcn}, Transformer, and LSTM, implemented with the tsai time-series library where applicable~\cite{tsai}. InceptionTime achieved the highest mean held-out-fold AUROC among the evaluated whole-night raw-sequence candidates and was selected as the D0 representative. The reported InceptionTime configuration uses a five-channel input, two output classes, batch size 8, Adam optimizer with a constant learning rate of $10^{-3}$, and cross-entropy loss with fold-specific class weights derived exclusively from training-fold labels. After 20 training epochs, the final-epoch model produces the held-out OOF probabilities.

\textbf{M0 configuration.} M0 partitions each night into overlapping 90-minute windows with 45-minute stride and aggregates valid window embeddings with gated attention pooling. Supplementary Section~\ref{app:m0_window_sensitivity} reports the window/stride sensitivity analysis that informed this windowing choice. The backbone family includes Transformer~\cite{phan2022sleeptransformer}, LSTM, ResNet, TCN~\cite{bai2018tcn}, and InceptionTime~\cite{ismail2020inceptiontime} under the five-channel input set with train-fold normalization. The primary M0 model uses a Transformer window encoder with input projection from five channels to hidden dimension 32, two Transformer encoder layers, four attention heads, GELU activation, feed-forward dimension 128, dropout 0.3, and output embedding dimension 64. The night-level classifier maps the pooled 64-dimensional embedding through a 32-unit hidden layer with ReLU and dropout to one logit. The reported Transformer MIL configuration uses batch size 8 bags, AdamW optimizer with a constant learning rate of $10^{-3}$ and weight decay $10^{-3}$, and BCE-with-logits loss with fold-specific positive-class weighting computed exclusively from training-fold labels. After 20 training epochs, the final-epoch model produces the held-out OOF probabilities and attention weights.

\section{Model-Family and Configuration Screening}
\label{app:model_families}

Exploratory candidate-model comparisons on the same patient-grouped folds informed the selection of one representative model for each paradigm before the shared pooled OOF analysis. Each comparison metric is computed separately on every held-out fold and summarized by its arithmetic mean and sample SD across the five folds. For B0 and B1, representative classifiers were selected within the prespecified SHAP-FS configuration. Tuned and default configurations without SHAP-FS are shown as sensitivity references. This differs from the main-manuscript results table, where held-out predictions are pooled before calculating each point estimate and patient-level bootstrap resampling provides the uncertainty interval. The configuration comparisons are exploratory because candidate selection and performance estimation used the same patient-grouped folds.

\begin{table*}[!htbp]
\centering
\caption{Exploratory configuration comparisons on the same patient-grouped folds. Values are mean $\pm$ sample SD of the five fold-specific held-out metrics and are not pooled OOF estimates. Only the selected SHAP-FS classifier is shown for B0 and B1. Tuned and default rows are non-SHAP-FS sensitivity references. D0 and M0 rows compare neural encoder families under the selected five-channel input set.}
\label{tab:appendix_model_families}
\scriptsize
\setlength{\tabcolsep}{2.0pt}
\begin{tabular}{llcccccc}
\toprule
Paradigm & Configuration / model & AUROC & AUPRC & F1 & Accuracy & Sensitivity & Specificity \\
\midrule
B0 & SHAP-FS Random Forest & 0.578 $\pm$ 0.103 & 0.798 $\pm$ 0.085 & 0.823 $\pm$ 0.061 & 0.709 $\pm$ 0.083 & 0.947 $\pm$ 0.058 & 0.080 $\pm$ 0.037 \\
B0 & Tuned Random Forest & 0.589 $\pm$ 0.080 & 0.811 $\pm$ 0.062 & 0.827 $\pm$ 0.049 & 0.714 $\pm$ 0.072 & 0.953 $\pm$ 0.027 & 0.087 $\pm$ 0.078 \\
B0 & Default Random Forest & 0.575 $\pm$ 0.083 & 0.806 $\pm$ 0.061 & 0.829 $\pm$ 0.048 & 0.714 $\pm$ 0.068 & 0.969 $\pm$ 0.026 & 0.040 $\pm$ 0.068 \\
\midrule
B1 & SHAP-FS Logistic Regression & 0.612 $\pm$ 0.056 & 0.819 $\pm$ 0.057 & 0.701 $\pm$ 0.091 & 0.617 $\pm$ 0.080 & 0.638 $\pm$ 0.118 & 0.535 $\pm$ 0.168 \\
B1 & Default XGBoost & 0.601 $\pm$ 0.083 & 0.816 $\pm$ 0.070 & 0.777 $\pm$ 0.049 & 0.656 $\pm$ 0.059 & 0.832 $\pm$ 0.035 & 0.191 $\pm$ 0.051 \\
B1 & Tuned XGBoost & 0.568 $\pm$ 0.058 & 0.798 $\pm$ 0.046 & 0.751 $\pm$ 0.066 & 0.625 $\pm$ 0.079 & 0.793 $\pm$ 0.066 & 0.174 $\pm$ 0.095 \\
\midrule
D0 & InceptionTime & 0.700 $\pm$ 0.075 & 0.859 $\pm$ 0.057 & 0.723 $\pm$ 0.092 & 0.655 $\pm$ 0.073 & 0.646 $\pm$ 0.133 & 0.671 $\pm$ 0.162 \\
D0 & TCN & 0.694 $\pm$ 0.056 & 0.859 $\pm$ 0.063 & 0.797 $\pm$ 0.061 & 0.694 $\pm$ 0.058 & 0.855 $\pm$ 0.130 & 0.240 $\pm$ 0.236 \\
D0 & Transformer & 0.664 $\pm$ 0.047 & 0.836 $\pm$ 0.035 & 0.706 $\pm$ 0.079 & 0.630 $\pm$ 0.060 & 0.630 $\pm$ 0.106 & 0.605 $\pm$ 0.152 \\
D0 & ResNet1D & 0.680 $\pm$ 0.059 & 0.857 $\pm$ 0.045 & 0.744 $\pm$ 0.044 & 0.655 $\pm$ 0.047 & 0.698 $\pm$ 0.072 & 0.543 $\pm$ 0.103 \\
D0 & LSTM & 0.543 $\pm$ 0.062 & 0.746 $\pm$ 0.029 & 0.700 $\pm$ 0.264 & 0.624 $\pm$ 0.161 & 0.787 $\pm$ 0.367 & 0.206 $\pm$ 0.387 \\
\midrule
M0 & Transformer & 0.698 $\pm$ 0.097 & 0.851 $\pm$ 0.072 & 0.745 $\pm$ 0.028 & 0.666 $\pm$ 0.025 & 0.681 $\pm$ 0.076 & 0.623 $\pm$ 0.222 \\
M0 & LSTM & 0.677 $\pm$ 0.084 & 0.841 $\pm$ 0.056 & 0.676 $\pm$ 0.080 & 0.606 $\pm$ 0.066 & 0.582 $\pm$ 0.100 & 0.681 $\pm$ 0.132 \\
M0 & ResNet & 0.669 $\pm$ 0.087 & 0.839 $\pm$ 0.059 & 0.580 $\pm$ 0.235 & 0.548 $\pm$ 0.149 & 0.501 $\pm$ 0.261 & 0.723 $\pm$ 0.305 \\
M0 & TCN & 0.651 $\pm$ 0.066 & 0.834 $\pm$ 0.041 & 0.770 $\pm$ 0.097 & 0.676 $\pm$ 0.087 & 0.781 $\pm$ 0.155 & 0.397 $\pm$ 0.230 \\
M0 & InceptionTime & 0.596 $\pm$ 0.057 & 0.803 $\pm$ 0.092 & 0.677 $\pm$ 0.144 & 0.593 $\pm$ 0.096 & 0.657 $\pm$ 0.261 & 0.450 $\pm$ 0.338 \\
\bottomrule
\end{tabular}
\end{table*}

Table~\ref{tab:appendix_model_families} provides the empirical basis for selecting one representative within each paradigm. Within the input-dimension-standardized SHAP-FS configurations, Random Forest and Logistic Regression had the highest mean AUROC for B0 and B1, respectively. The non-SHAP-FS tuned Random Forest had a marginally higher mean AUROC than B0 SHAP-FS Random Forest but similarly weak specificity, and is reported only as a configuration sensitivity reference. InceptionTime had the highest mean AUROC among the evaluated D0 candidates under the shared class-imbalance protocol and maintained substantially greater specificity than TCN, whose high sensitivity was accompanied by low and variable specificity. The 90-minute M0 backbone comparison favored the Transformer encoder, which was selected as the MIL representative.

Interpretation of the candidate-model comparisons is constrained by the small cohort, reuse of the same grouped folds for candidate selection and final evaluation, and high event prevalence, under which thresholded metrics can appear favorable despite weak specificity. Variants beyond the four selected models are therefore presented as exploratory configuration evidence rather than as additional primary comparisons.

\section{M0 Window and Stride Sensitivity}
\label{app:m0_window_sensitivity}

Table~\ref{tab:m0_window_sensitivity} summarizes the Transformer MIL window/stride sensitivity analysis under the five-channel, train-fold-normalized input set. These values are simple five-fold means used to select the M0 windowing configuration. The 90-minute window with a 45-minute stride had the highest mean AUROC and AUPRC among the evaluated settings, while the 180-minute setting was competitive but provided coarser temporal resolution.

\begin{table}[!htbp]
\centering
\caption{M0 Transformer window/stride sensitivity under five-channel train-fold normalization. Values are simple five-fold means for sensitivity screening.}
\label{tab:m0_window_sensitivity}
\footnotesize
\setlength{\tabcolsep}{3.5pt}
\begin{tabular}{lcccccc}
\toprule
Window / stride & AUROC & AUPRC & F1 & Accuracy & Sensitivity & Specificity \\
\midrule
30 / 15 min & 0.622 & 0.805 & 0.718 & 0.624 & 0.668 & 0.518 \\
60 / 30 min & 0.645 & 0.826 & 0.719 & 0.630 & 0.668 & 0.541 \\
90 / 45 min & 0.698 & 0.851 & 0.745 & 0.666 & 0.681 & 0.623 \\
180 / 90 min & 0.677 & 0.837 & 0.703 & 0.624 & 0.635 & 0.621 \\
\bottomrule
\end{tabular}
\end{table}

\section{Higher-Burden Agitation Endpoint Sensitivity}
\label{app:severe_gt4}

The primary endpoint treats any PAS domain score above zero as next-day agitation. PAS rates four agitation domains individually from 0 to 4~\cite{rosen1994pas}. As a stricter composite endpoint, Table~\ref{tab:severe_gt4} defines higher-burden agitation as a day-level PAS total score greater than 4, calculated as the sum of the maximum recorded motor, verbal, aggression, and resistance-to-care domain scores. This threshold requires agitation in at least two domains and at least one domain score above 1. The same 423 patient-nights were included, but the number of positive patient-nights decreased to 73 nights (17.3\%). Under this lower-prevalence endpoint, D0 retained the highest AUROC, AUPRC, sensitivity, and balanced accuracy, whereas B1 achieved the lowest Brier score and highest specificity. The sensitivity analysis shows partial robustness of the D0 ranking under a lower-prevalence, higher-burden endpoint, while demonstrating that model operating characteristics and the practical interpretation of predictions depend strongly on the agitation definition.

\begin{table}[htbp]
\centering
\caption{Higher-burden agitation endpoint sensitivity using a day-level PAS total score \(>4\) as the positive class. Metrics are pooled OOF point estimates on 423 patient-nights with 17.3\% prevalence.}
\label{tab:severe_gt4}
\footnotesize
\setlength{\tabcolsep}{4.0pt}
\begin{tabular}{lcccccc}
\toprule
Model & AUROC & AUPRC & Brier & Sens. & Spec. & BalAcc \\
\midrule
B0 & 0.566 & 0.255 & 0.178 & 0.178 & 0.900 & 0.539 \\
B1 & 0.636 & 0.280 & 0.164 & 0.260 & 0.911 & 0.586 \\
D0 & 0.654 & 0.286 & 0.239 & 0.575 & 0.700 & 0.638 \\
M0 & 0.635 & 0.267 & 0.179 & 0.274 & 0.877 & 0.576 \\
\bottomrule
\end{tabular}
\end{table}

\section{Per-Patient Heterogeneity}
\label{app:per_patient}

Within-patient discrimination was characterized for all four paradigms to distinguish pooled cross-patient performance from variation in patient-level ranking. Per-patient AUROC is defined only when a patient has at least one agitated and one non-agitated night. In this cohort, 37 of 65 patients met that condition, while 28 patients had single-class outcomes and therefore undefined within-patient AUROC. Among the 37 evaluable patients, patient-level time series remained short, with a median of 7 nights and a range of 3--10 nights per patient.

\begin{figure}[htbp]
\centering
\includegraphics[width=0.5\linewidth]{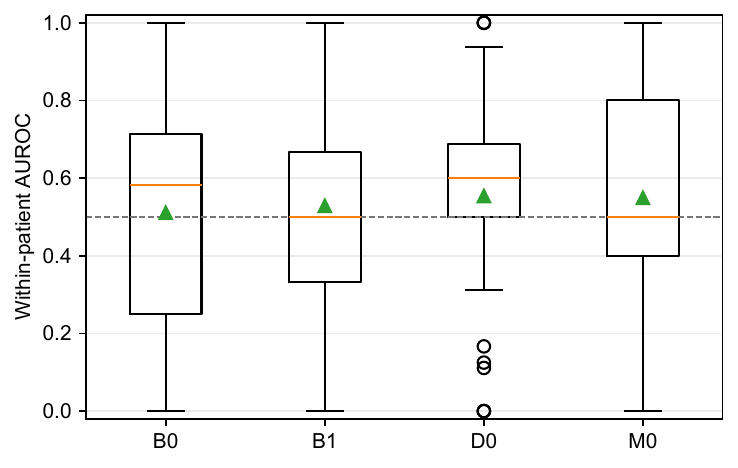}
\caption{Within-patient AUROC by paradigm. AUROC is computed only for patients with at least one agitated and one non-agitated night. Of the 65 patients, 37 were evaluable and 28 with single-class outcomes were excluded. Lines within boxes denote medians, green triangles denote means, and the dashed line marks chance-level AUROC=0.50.}
\label{fig:per_patient_boxplot}
\end{figure}

Figure~\ref{fig:per_patient_boxplot} summarizes the resulting distribution. The horizontal line inside each box denotes the median patient-level AUROC, the green triangle denotes the mean, and the dashed horizontal line marks chance-level within-patient discrimination (AUROC=0.50). D0 had the highest median within-patient AUROC (0.60), followed by B0 (0.58), while B1 and M0 had medians of 0.50. The broad interquartile ranges and outlying low-AUROC patients indicate substantial heterogeneity, and within-patient discrimination remains uncertain over the available short follow-up.

\section{Single-Subject Prediction Trajectories}
\label{app:single_subject}

Patients with at least seven observed nights were eligible for illustration. The outcome-varying examples were selected deterministically as the patients with the largest numbers of non-agitated and agitated nights, respectively. The single-class examples were selected by longest follow-up, with ties resolved by greater cross-model prediction variability. These panels illustrate predefined patient-level behavior patterns and are not used for inferential comparisons. Subject-level trajectories use OOF probabilities to remain consistent with the held-out-patient primary evaluation and provide qualitative context for the heterogeneity summarized in Fig.~\ref{fig:per_patient_boxplot}. For the two outcome-varying patients, M0 generally assigned higher probabilities to agitation-positive than agitation-negative nights, whereas D0 showed limited separation. The single-class examples further demonstrate differences in absolute risk assignment: B0 consistently assigned high probabilities to the all-negative patient, while D0 frequently assigned probabilities below 0.5 to the all-positive patient. These examples illustrate that pooled discrimination can coexist with an inability to rank agitated above non-agitated nights within some patients, systematic risk-level differences, and unstable fixed-threshold classifications. The selected panels characterize possible model behaviors rather than estimate their prevalence across the cohort. The colored date labels mark the next-day agitation outcome associated with each preceding sensor night.

\begin{figure}[htbp]
\centering
\begin{minipage}{0.49\linewidth}
\centering
\includegraphics[width=0.96\linewidth]{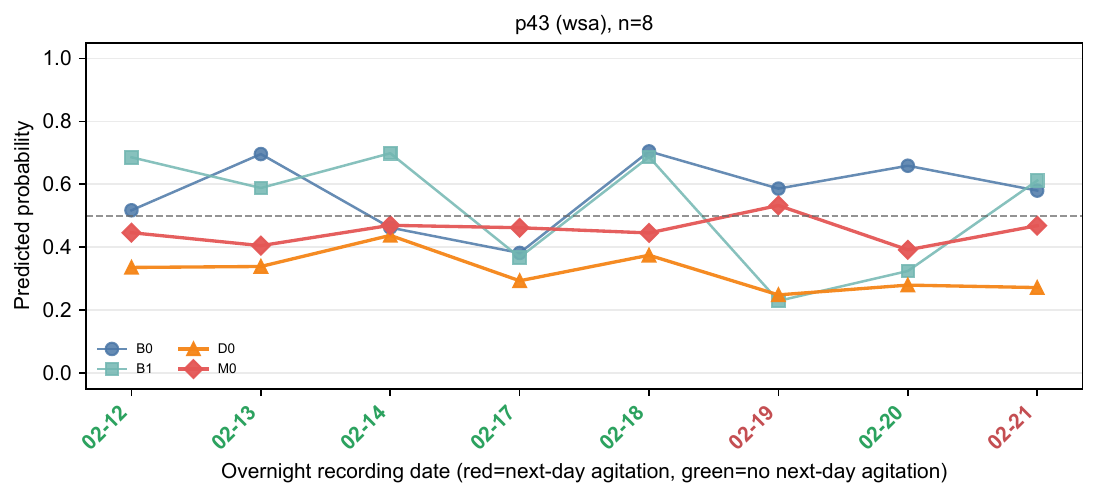}
\end{minipage}
\hfill
\begin{minipage}{0.49\linewidth}
\centering
\includegraphics[width=0.96\linewidth]{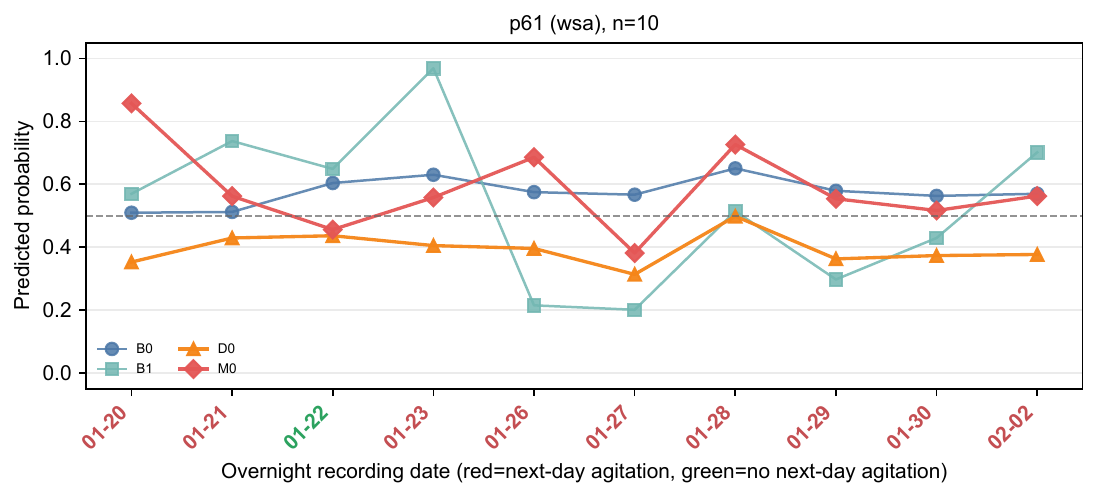}
\end{minipage}

\vspace{0.2em}

\begin{minipage}{0.49\linewidth}
\centering
\includegraphics[width=0.96\linewidth]{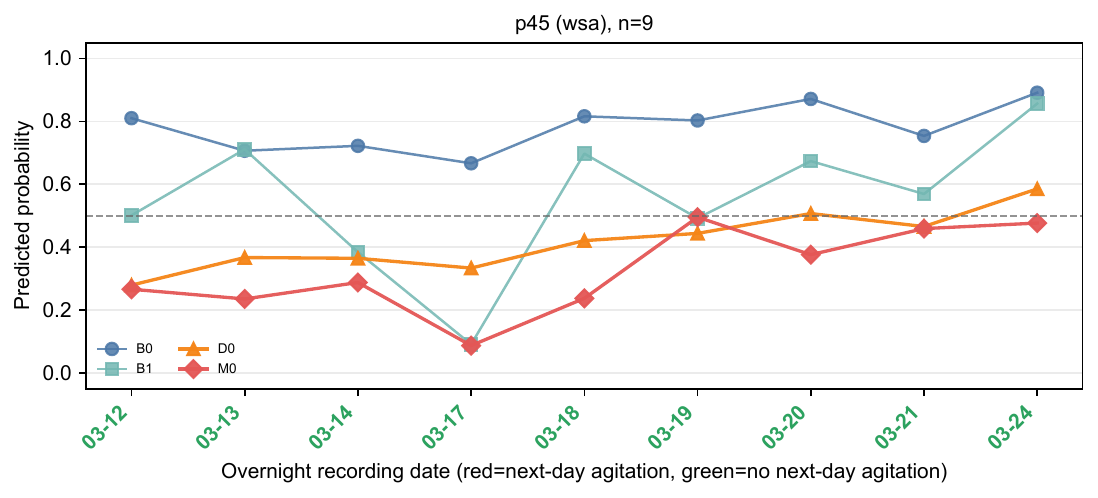}
\end{minipage}
\hfill
\begin{minipage}{0.49\linewidth}
\centering
\includegraphics[width=0.96\linewidth]{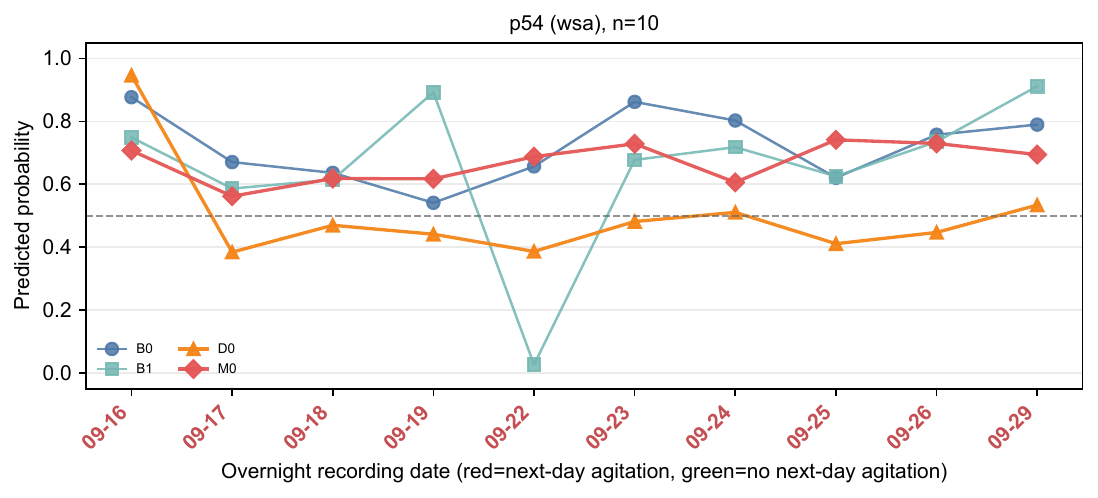}
\end{minipage}
\caption{Selected single-subject prediction trajectories from pooled OOF predictions. Patients with at least seven observed nights were eligible. The outcome-varying panels were selected by the largest numbers of non-agitated and agitated nights, respectively. The single-class panels were selected by longest follow-up, with ties resolved by greater cross-model prediction variability. Red and green date labels indicate agitated and non-agitated next-day outcomes, respectively. Each point is predicted from the preceding overnight sensor record.}
\label{fig:single_subject_examples}
\end{figure}

\end{document}